\documentclass[conference]{IEEEtran}
\usepackage{cite}
\ifCLASSINFOpdf
\else
\fi
\usepackage[T1]{fontenc}
\usepackage{graphicx}
\usepackage{hyperref}
\usepackage{lineno}
\usepackage{graphicx}
\usepackage{enumerate}
\begin{document}
%

%
\title{Discrete Wavelet Transform and Gradient Difference based approach for text localization in videos}
\author{\IEEEauthorblockN{B.H.Shekar}
\IEEEauthorblockA{Department of Computer Science,\\ Mangalore University,\\Mangalore, Karnataka,\\ India.\\
Email: bhshekar@gmail.com}
\and
\IEEEauthorblockN{Smitha M.L.}
\IEEEauthorblockA{Department of Master of Computer Applications,\\ 
KVG College of Engineering,\\Sullia,Karnataka,\\India.\\
 Email: smithaml.urubail@gmail.com}
\and
\IEEEauthorblockN{P. Shivakumara}
\IEEEauthorblockA{Department of CSIT,\\University of Malaya,\\Kaulalumpur,\\Malayasia\\
 Email: shiva@um.edu.my}
}
\maketitle
\begin{abstract}
The text detection and localization is important for video analysis and understanding. The scene text in video contains semantic information and thus can contribute significantly to video retrieval and understanding. However, most of the approaches detect scene text in still images or single video frame. Videos differ from images in temporal redundancy. This paper proposes a novel hybrid method to robustly localize the texts in natural scene images and videos based on fusion of discrete wavelet transform and gradient difference. A set of rules and geometric properties have been devised to localize the actual text regions. Then, morphological operation is performed to generate the text regions and finally the connected component analysis is employed to localize the text in a video frame. The experimental results obtained on publicly available standard ICDAR 2003 and Hua dataset illustrate that the proposed method can accurately detect and localize texts of various sizes, fonts and colors. The experimentation on huge collection of video databases reveal the suitability of the proposed method to video databases.
\end{abstract}
%
\IEEEpeerreviewmaketitle

\section{Introduction}
The text embedded in images and videos contain lots of semantic information which are useful for video comprehension. In recent years, the automatic detection of texts from images and videos has gained increasing attention. However, the large variations in text fonts, colors, styles, and sizes, as well as the low contrast between the text and the complicated background often make text detection extremely challenging.	

In general, the methods for detecting text can be broadly categorized into three groups: connected component based (CC), edge based and texture based methods.  The CC-based methods apply a bottom-up approach by grouping small components into successively larger ones until all regions are identified in the image. However, these CCs may not preserve the full shape of the characters due to color bleeding and the low contrast of the text lines. Therefore, these methods do not work well for video frames because it assumes that text pixels in the same region have similar colors or grayscale intensities.

To overcome the problem of low contrast, edge-based methods are proposed. Zhang et al.~\cite{zhang} proposed to extract statistical features from the Sobel edge maps of four directions and  K-means was used to classify pixels into the text and nontext clusters. Although this method is robust against complex background, it fails to detect low contrast text and texts of small font sizes. It is also computationally expensive due to the large feature set. Wong et al.~\cite{Wong2003} computed maximum gradient difference  to identify the potential line segments. This method has a low false positive rate, but it uses many heuristic rules and is sensitive to threshold values. Liu et al.~\cite{Liu_05} used edge-based methods that are fast, but they produced many false positives for images with complex backgrounds. The edge based approach requires text to have a reasonably high contrast to the background in order to detect the edges. So, these methods often encounter problems with complex backgrounds and produce many false positives.

To overcome the problem of complex background, the texture based approaches consider text as a special texture. These methods apply transform based principle such as fast fourier transform, discrete cosine transform~\cite{HuangX}, wavelet decomposition~\cite{Shivkum}, and gabor filters for feature extraction. Texture-based methods~\cite{Chen},~\cite{Ye:2005} scan the image at a number of scales, classify neighborhoods of pixels based on number of text properties, such as high density of edges, low gradients above and below threshold, high variance of intensity, distribution of wavelet or DCT coefficients, etc.  However, these methods require extensive training and are computationally expensive for large databases. Additionally, these algorithms are typically unable to detect the slanted text. 

Wavelet decomposition generally enhances the high contrast pixels by suppressing low contrast pixels. Ye et al.~\cite{Ye:2005} computed  wavelet energy features at different scales and  employed adaptive thresholding to find candidate text pixels, which are then merged into candidate text lines. The multiple frames integration approach~\cite{YenShwu} was used to detect and localize static caption texts on news videos. Wei et al.~\cite{Weicheng} deployed pryramidal scheme where two downsized images are generated by bilinear interpolation. Then, the gradient difference~\cite{Trung} of each pixel is calculated for three differently sized images, including the original one. Finally, text candidates were identified through verification phase. In the first phase, the  geometrical properties and texture of each text candidate are obtained. In the second phase, statistical characteristics of the text candidate are computed using a discrete wavelet transform, and then the principal component analysis is used to reduce the number of dimensions of these features. Next, the optimal decision function of the support vector machine, obtained by sequential minimal optimization was applied to determine whether the probable text regions contain texts or not. Wang et al.~\cite{WangJain} used wavelet transform to extract features that represent the texture property of text regions and an unsupervised fuzzy c-means classifier was used to discriminate candidate text pixels from background. 

From our survey, we have noticed that most of the works focus on the detection of horizontal text in still images but not multi-oriented text. This is because, most of the non-horizontal text lines are scene text, which is much more difficult to detect due to varying lighting and complex transformations~\cite{jung},~\cite{zhang}. In addition, most of the proposed methods discussed the experimental works on static images, but not on the video frames~\cite{WangB}. In this context, we propose a video based text localization system that takes into account, shot detection from a video, key frame identification from the shots followed by text localization in each key frames.

The remaining part of the paper is organized as follows. The proposed approach is discussed in section 2. Experimental results and comparison with other approaches are presented in section 3 and conclusion is provided in section 4.
\section{Proposed Methodology}
\begin{figure}[htb]
 \begin{center}
\includegraphics[scale=0.3]{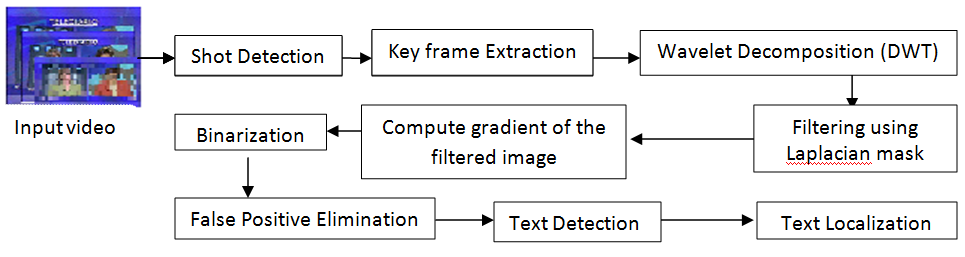}
\caption{Flowchart of the proposed system} 
\label{fig1}
 \end{center}
\end{figure}

The flowchart of the proposed text detection and localization approach is shown in Fig.~\ref{fig1}. The details of each processing blocks are discussed below.
\subsection{Shot detection}
In this section, we present the proposed approach that extract keyframes based on color moments for subsequent processing. Given an input video that contains many frames, we compute the colour moments for each frame and the Euclidean distance measure is used to measure the similarity between the frames. Based on the set threshold, a shot is said to be detected if the dissimilarity between the frames is very high. From each shot, a key frame is extracted based on spatio-temporal color distribution~\cite{Shekar}.
The color distributions of the $ Y $,$ I $ and $ Q $ components of a frame is represented by its color moments and probability distribution is uniquely characterized by its moments. We compute the color moments of a frame to capture the characteristics of a frame as follows.
The first color moments of the $ i^{th} $ color component is given by
 \begin{equation}
M_{i}^{'} = \frac{1}{N}\sum_{k=1}^{N}  p_{i,k} \ , \ \ i=1,2,3.
 \end{equation} 

where $ p_{i,k} $ is the intensity value of the $ i^{th} $ color component of the $ k^{th} $  pixel of a frame and $ N $ is the total number of pixels in the frame.  \\
The $h^{th}$  moment, $h $ = 2,3, .., of the $i^{th}$ color component  is then given by
\begin{equation}
M_{i}^{h}=\left( \frac{1}{N}\sum_{k=1}^{N}\left( p_{i,k}-M_{i}^{'}\right) ^{h}\right) ^{\frac{1}{h}}
\end{equation}
Compute the first $ H $ moments for each color component in frame $f_{j}$ to form a feature vector, $F_{j}$, where  $j=1,2,..n,$ as follows:
 
\begin{eqnarray}
F_{j}=[ct_{1},ct_{2},...ct_{z}] \nonumber \\
= [ \alpha_{1}M_{1}^{1},\alpha_{1}M_{1}^{2},\alpha_{1}M_{1}^{3},... \alpha_{1}M
_{1}^{H}, \nonumber \\ \alpha_{2}M_{2}^{1},\alpha_{2}M_{2}^{2},\alpha_{2}M_{2}^{3},... \alpha_{2}M_{2}^{H},   \\
\alpha_{3}M_{3}^{1},\alpha_{3}M_{3}^{2},\alpha_{3}M_{3}^{3}, ... \alpha_{3}M_{3}^{H} ]  \nonumber
\end{eqnarray}

Here  $z = 3 $  and    are the weights for the $Y$,$I$,$Q$ components. Then the difference between the frames  $ f{_{j}} $ and $ f{_{j-1}}$  is computed as follows:

 \begin{equation}
M_{i}^{h}=\left( \frac{1}{N}\sum_{k=1}^{N}\left( p_{i,k}-M_{i}^{'}\right) ^{h}\right) ^{\frac{1}{h}}
\end{equation}

where  
\begin{equation}
d(f_{j},f_{j-1})=\sum_{k=1}^{z}\parallel F_{j}(k)-F_{j-1}(k)\parallel ^{q}
\end{equation}
When $ q $ is set to 2, $d(i, j)$ is the Euclidean distance between the frame $i$ and frame $ j $. The value of $D$ indicates the change tendency of consecutive frames. If $ |D| > T1 $, then a cut is detected, where $ T1 $ is the cut threshold~\cite{ZhangD}.	
\subsection{Key Frame Extraction}
The key frame extraction is based on spatio-temporal color distribution~\cite{SunChen}. Firstly, we construct a temporally maximum occurrence frame which considers the spatial and temporal distribution of the pixels throughout the video shot. Then, a weighted distance is computed between frames in the shot and the constructed reference frame. The key frames are extracted at the peaks of the distance curve and can achieve high compression ratio and high fidelity.
\subsection{Edge map extraction using Discrete Wavelet Transform}
The discrete wavelet transform(DWT)~\cite{Weicheng} is a very useful tool for signal analysis and image processing, especially in multi-resolution representation. In the field of signal analysis and image processing, the DWT is very useful tool that can decompose the signal into different components in the frequency domain. In case of 1-D DWT, it decomposes signal into two components: one is average and another one is detail component. The 2-D DWT decomposes an image into four components or sub-bands, one average component(LL) and three detail components(LL, HL, HH). This 2-D DWT finds its application in image processing domain including edge detection. The main reason of applying wavelet transform for edge detection is that wavelet transform can remove the noise whereas conventional edge operators identifies noisy pixels as edge pixels. In our proposed algorithm, we are using 2-D multilevel discrete wavelet transform which provides a powerful tool for modeling the characteristics of text-like images.  
\subsection{Gradient Difference based Text Localization}
The text-like image obtained by the application of multilevel DWT is further processed to localize the actual text blocks.We propose to employ the Laplacian mask on the DWT bands to identify the text regions. Since the mask produces two values for every edge, the Laplacian-filtered image contains both positive and negative values. The transitions between these values (the zero crossings) correspond to the transitions between text and background.  The result obtained in the previous step is now filtered using $3 \times 3$ Laplacian mask. In order to capture the relationship between positive and negative values, we use the maximum gradient difference(MGD)~\cite{Trung}, defined as the difference between the maximum and minimum values within a local  $1 \times N$ window.The MGD value at pixel $(i, j)$ is computed from the Laplacian-filtered image $f$ as follows.

\begin{figure}
\centering
\includegraphics[scale=0.3]{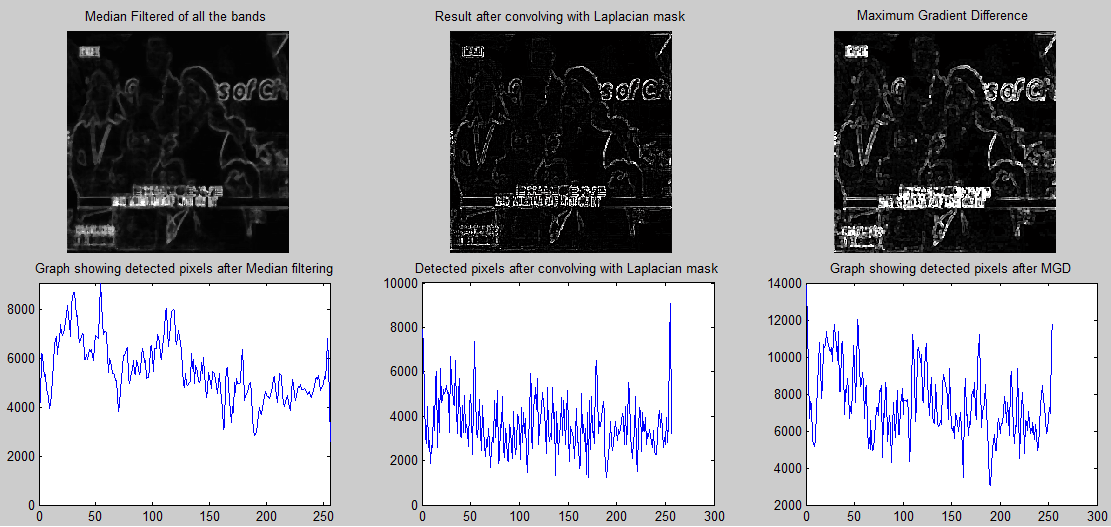} 
\caption{Intermediate results to obtain MGD}
\label{fig2}
\end{figure}

Let $f$ be the Laplacian filtered image and GD be the gradient difference image defined as:\\
   $GD(x,y) = Max(f(x,y)) - Min(f(x,y))$ \\
The MGD map is obtained by moving the window over the image. The brighter region represent larger MGD values which corresponds to the text regions as shown in Fig.~\ref{fig2}. 
\subsection{False Positive Elimination}
\begin{figure}
\centering
\includegraphics[scale=0.3]{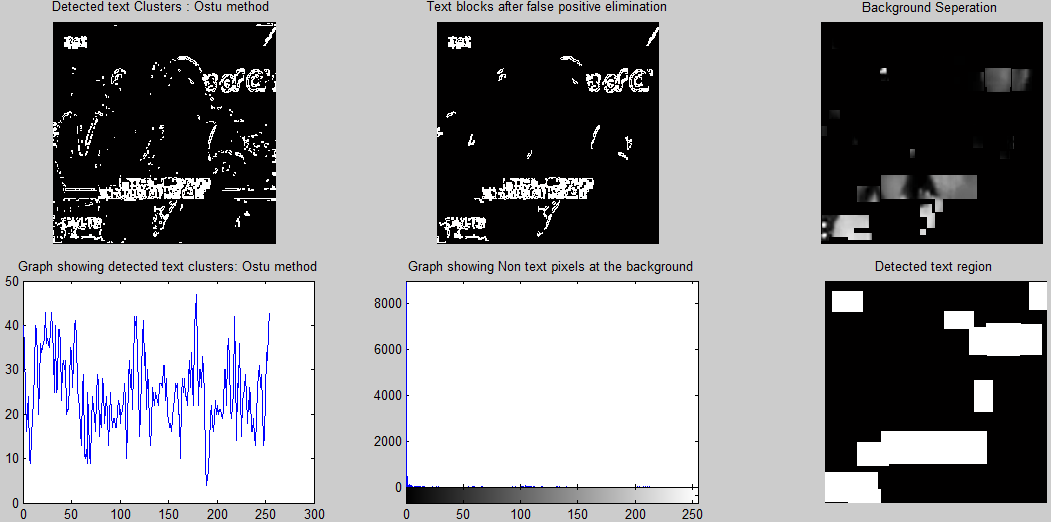} 
\caption{Graph showing detected text and non text pixels}
\label{fig3}
\end{figure}

The significant text region obtained due to Laplacian filter may also contain non-text blocks. Hence, we propose to employ some geometrical rules to eliminate the non-text regions. In this false elimination process~\cite{Lyum}, we perform binarization by using Ostu method to obtain the text clusters. In order to filter out the non-text objects, some of the geometric features are computed. The false positives are eliminated by computing height, width, aspect ratio$(AR)$ and using some geometrical rules devised based on edge area$(EA)$ of the text blocks. 
         \begin{center}
          $ AR = width / height$ 
          \end{center}
             \begin{center}
              $ density = EA / ( height * width ) $
              \end{center}  
According to the attributes of the horizontal text line, we make the following rules to confirm on the non text blocks.
       \begin{center}
        $  i)  AR < T1  ||   density < T2  $
        \end{center}
         \begin{center}
        $  ii) height > 50 ||  height < 6  $
        \end{center}
         \begin{center}
        $ iii) width < 5 ||  height * width < 24 $ 
        \end{center}
By these rules, we can get the candidate text lines. Then, we label the connected components by using 4-connectivity. The foreground connected components for each of these frames are considered as text candidates. The thresholding values $T1$ and $T2$ are the mean and standard deviation of the pixel values respectively. Then, morphological dilation is performed to fill the gaps inside the obtained text regions which yields better results and the boundaries of text regions are identified as shown in Fig.~\ref{fig3}. 
\section{Experimental Results}
This section present the experimental results to reveal the success of the proposed approach. To exhibit the performance of our system we carried out exhaustive evaluations using large collection of videos, collected from the internet. In addition, to present the suitability of the proposed approach even for document images, we have conducted experimentation on ICDAR-2003 and Hua$'$s dataset which are said to be the bench mark datasets and considered by many researchers to evaluate their approaches. We have also made a comparative analysis with some of the well known algorithms to exhibit the performance of the proposed approaches. 
\subsection{Dataset and Comparison Metrics}
The proposed approach is evaluated with respect to f-measure which is a combination of two measures:   precision and recall. It is observed that most of the text blocks of each video frame possess properties such as varying fonts, colour, size, languages etc. The detected text blocks in an image are represented by their bounding boxes. We evaluate the performance at the block level which is a common granularity level presented in the literature, eg.,~\cite {Shivakumara},~\cite{Shivkum} rather than the word or character level. To judge the correctness of the text blocks detected, we manually count the true text blocks present in the frame.  Further, we manually label each of the detected blocks as one of the following categories.

The \textit{truly detected text block}($TDB$) is a detected block that contains partially or fully text. The \textit{falsely detected text block}($FDB$) is a block with false detections. The \textit{text block with missing data}($MDB$) is a detected text block that misses some characters. Based on the number of blocks in each of the categories mentioned above, the following metrics are calculated to evaluate the performance of the method. 
 \begin{center}
 Detection rate = Number of TDB / Actual number of text blocks \\
 False positive rate=Number of FDB / Number of (TDB + FDB) \\
 Misdetection rate = Number of MDB / Number of TDB
 \end{center}
\begin{figure}
\centering
\includegraphics[scale=0.3]{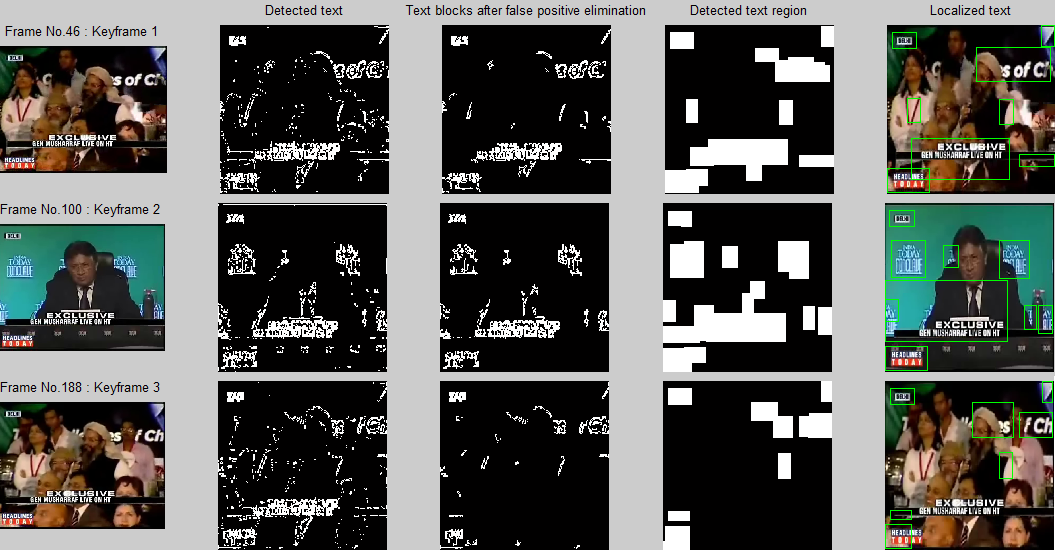} 
\caption{Sample results of video text localization}
\label{fig4}
\end{figure}
\begin{table}
\centering
\caption{Evaluation performance : Precision and Recall metrics to some sample videos }
\begin{tabular}{|p{1.1cm}|p{0.5cm}|p{0.4cm}|p{0.6cm}|p{0.7cm}|p{0.9cm}|p{0.4cm}|}
\hline 
Test Videos& Key frames& ATB & Recall & Precision & F measure & MDR \\ 
\hline 
English News & 3 & 18 & 90 & 31 & 41 & 25 \\ 
\hline 
Kannada News & 2 & 6 & 100 & 50 & 80 & 17 \\ 
\hline 
Chinese News & 1  & 2 & 100 & 33 & 50 & 0 \\ 
\hline 
Sports Clip & 2 & 6 & 100 & 50 & 67 & 0 \\ 
\hline 
Cartoon Show & 1 & 1 & 100 & 100 & 0 & 0 \\ 
\hline 
Chinese Lecture & 2 & 1 & 100 & 50 & 67 & 0 \\
\hline 
Motivational Video & 3 & 6 & 100 & 63 & 62 & 0 \\ 
\hline 
\end{tabular}  
\label{tab1}
\end{table}
In Fig.~\ref{fig4}, we have shown the experimental results that we have obtained on a sample video at each stage. There are three key frames in this sample video, where each key frame respectively contain varied number of text blocks with varying font size and color. The performance of the proposed approach is evaluated for different videos using precision and recall as evaluation metrics and the results are reported in Table~\ref{tab1}. Experimentation was also performed on still images and the localized text blocks  are shown in Fig.~\ref{fig5}. It shall be observed that the proposed approach works effectively for images with slant text/ non-horizontal text and the curvy text. \\
\begin{figure}
\includegraphics[scale=0.3]{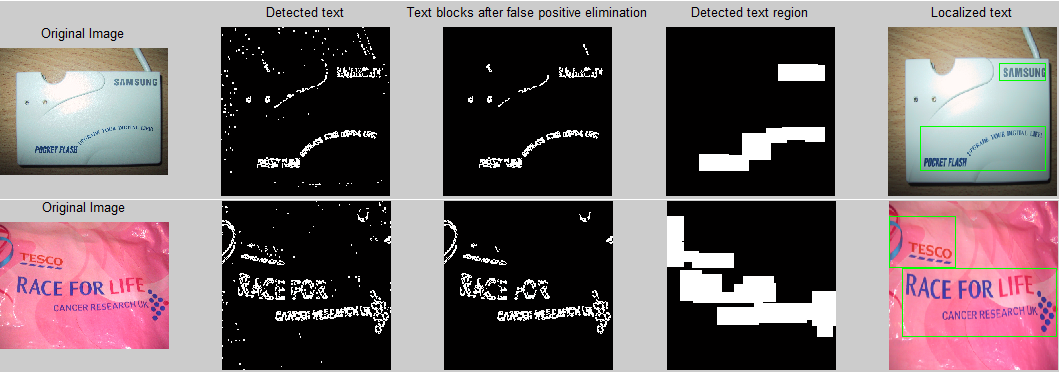}
\caption{Sample results of still image localization}
\label{fig5}
\end{figure}
The performance of the proposed approach is evaluated for still images using precision and recall as evaluation metrics and   the results are reported in Table~\ref{tab2}, ~\ref{tab3} and ~\ref{tab4} respectively. In order to exhibit the performance of the proposed approach, we have made a comparative study with the state of the art text localization approaches. We have chosen gradient based method~\cite {Shivakumara} and wavelet based method~\cite{Shivkum} for comparative analysis as these methods are proved to be the most suitable works for text localization in document images. The gradient based method~\cite {Shivakumara} basically suffers from several thresholds for identifying text segments and hence the performance degrades compared to the proposed method as it is user dependent. The wavelet based method~\cite{Shivkum} gives better results because of the advantages of wavelet and color features for text enhancement. Based on experimental results on ICDAR-03 , Horzontal and  Hua's dataset, it is concluded that discriminating false positives and true text blocks is not easy. However, this method is found to be suitable for document images and the performance of this approach degrade when tested on video frames because of the inherent complexity that exists in video frames. Tables ~\ref{tab2}, ~\ref{tab3} and ~\ref{tab4} shows that the performance of the proposed method is better than the existing methods for ICDAR-03, Hua data and Horizontal data images. Table~\ref{tab1} illustrates the performance evaluation on sample videos and the results are much better when compared to the existing approaches. Thus, the proposed method has high detection rate and low false positive rate than the existing methods and hence is suitable for both document images and video databases.  
\begin{table}
\centering
\caption{Evaluation performance : Precision and Recall metrics for ICDAR-03 images }
\begin{tabular}{|c|c|c|c|c|}
\hline 
METHODS & R & P & F & MDR \\ 
\hline 
 Gradient Based~\cite {Shivakumara} & 51.6 & 16.5 & 25.0 & 8.2  \\ 
\hline 
Wavelet Based~\cite{Shivkum} & 54.0 & 16.4 & 8.4 & 6.5  \\ 
\hline 
Proposed & 94.6 & 83.8 & 89.5 & 0.03  \\ 
\hline 
\end{tabular} 
\label{tab2}
\end{table}
\begin{table}
\centering
\caption{Evaluation performance : Precision and Recall metrics for HUA data }
\begin{tabular}{|c|c|c|c|c|}
\hline 
METHODS & R & P & F & MDR \\ 
\hline 
 Gradient Based~\cite {Shivakumara} & 51.6 & 16.5 & 25.0 & 8.2  \\ 
\hline 
Wavelet Based~\cite{Shivkum} & 54.0 & 16.4 & 8.4 & 6.5  \\ 
\hline 
Proposed & 95.7 & 85.6 & 90.0 & 0.4  \\ 
\hline 
\end{tabular} 
\label{tab3}
\end{table}
\begin{table}
\centering
\caption{Evaluation performance : Precision and Recall metrics for HORIZONTAL data }
\begin{tabular}{|c|c|c|c|c|}
\hline 
METHODS & R & P & F & MDR \\ 
\hline 
 Gradient Based~\cite {Shivakumara} & 51.6 & 16.5 & 25.0 & 8.2  \\ 
\hline 
Wavelet Based~\cite{Shivkum} & 54.0 & 16.4 & 8.4 & 6.5  \\ 
\hline 
Proposed & 92.98 & 44.23 & 60.0 & 4.0  \\ 
\hline 
\end{tabular} 
\label{tab4}
\end{table}
\section{Conclusion}
Video text contains abundant high-level semantic information which is important to video analysis, indexing and retrieval. The existing works were shown to be accurate for document images but not for video database because of inherent complexity hidden in a text(color,size,font,illumination,etc.) present in video frames. In this context, the newly developed hybrid approach is capable of localizing the text regions accurately both in video frames and document images. Experimental results and comparative study with existing methods have shown that the proposed method works for both videos and static images. 

\end{document}